\pdfoutput=1

\documentclass[11pt]{article}

\usepackage[]{acl}

\usepackage{times}
\usepackage{latexsym}

\usepackage{comment}
\usepackage{graphicx}
\usepackage{caption}
\usepackage{subcaption}
\usepackage{amsmath}
\usepackage{dsfont}
\usepackage{booktabs}
\usepackage{multirow}

\usepackage[T1]{fontenc}

\usepackage[utf8]{inputenc}

\usepackage{microtype}

\newcommand{\eg}{{\it e.g.}}
\newcommand{\ie}{{\it i.e.}}

\newcommand{\our}{EADPR}

\title{Evidentiality-aware Retrieval for Overcoming Abstractiveness in Open-Domain Question Answering}

\author{Yongho Song$^{1}$\thanks{~~Equal contribution}  \qquad \textbf{Dahyun Lee}$^{1*}$ \qquad Myungha Jang$^{1}$ \\
\textbf{Seung-won Hwang}$^{2}$  \qquad \textbf{Kyungjae Lee}$^{3}$ \qquad \textbf{Dongha Lee}$^{1}$ \qquad \textbf{Jinyoung Yeo}$^{1}$ \\ \\ Yonsei University$^{1}$  \quad Seoul National University$^{2}$ \quad LG AI Research$^{3}$ \\
\texttt{\{kopf\_yhs, leedhn, donalee, jinyeo\}@yonsei.ac.kr} \\
\texttt{myunghajang@gmail.com} \quad \texttt{seungwonh@snu.ac.kr} \\
\texttt{kyungjae.lee@lgresearch.ai} \\
}

\begin{document}

\maketitle

\begin{abstract}

The long-standing goal of dense retrievers in abtractive open-domain question answering (ODQA) tasks is to learn to capture evidence passages among relevant passages for any given query, such that the reader produce factually correct outputs from evidence passages. One of the key challenge is the insufficient amount of training data with the supervision of the answerability of the passages. Recent studies rely on iterative pipelines to annotate answerability using signals from the reader, but their high computational costs hamper practical applications. In this paper, we instead focus on a data-centric approach and propose Evidentiality-Aware Dense Passage Retrieval (EADPR), which leverages synthetic distractor samples to learn to discriminate evidence passages from distractors. We conduct extensive experiments to validate the effectiveness of our proposed method on multiple abstractive ODQA tasks. 

\end{abstract}

\section{Introduction}\label{sec:intro}

Information retrieval (IR) has served as a core component in open-domain question answering (ODQA)~\citep{kwiatkowski2019natural,joshi2017triviaqa}, which require the model to produce factually correct outputs based on a vast amount of knowledge in an unstructured text corpus. The predominant approach to ODQA employs the simple yet effective retriever-reader framework~\citep{chen2017reading}, where the retriever (\ie, IR system) finds contexts that are relevant to the query from a large collection of texts, and the reader infers the final answer from the retrieved contexts. While augmenting the reader with a retriever is helpful when answerability aligns well with the relevance from the retriever, such an assumption does not always hold in abstractive ODQA tasks, \eg, multi-hop QA~\citep{yang2018hotpotqa}, where target passages do not necessarily include the answer to the question.

\begin{figure}[t]
    \centering
    \includegraphics[width=\linewidth]{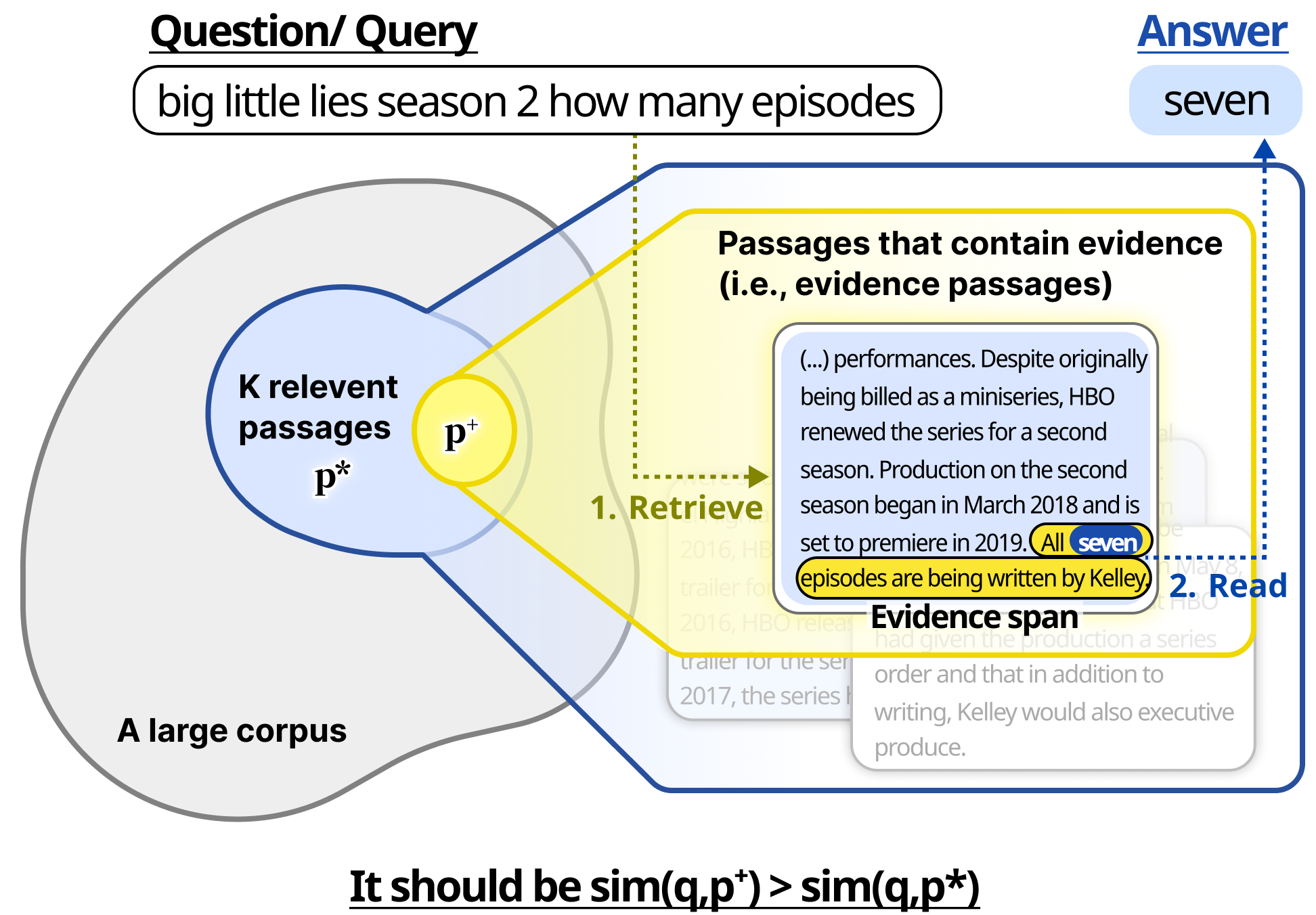}
    \caption{A bird's-eye view of the goal of passage retrieval in abstractive tasks. An ideal retriever (1) retrieves evidence passages such that the reader (2) produces answers based on the evidence span.}
    \label{fig:concept}
\end{figure}

The misalignment between relevance and answerability in abstractive tasks poses a significant challenge to IR systems. 
The standard approach to building an IR system leverages human-annotated pairs of questions and relevant passages~\citep{nguyen2016ms}, but these IR datasets based on relevance provide only a weak supervision signal to abstractive tasks. 
This is particularly crucial for state-of-the-art IR systems, which train a dense passage retriever (DPR)~\citep{karpukhin2020dpr} using the relevance annotations to find relevant passages for a given question based on their learned vector representations. Training a dense retriever with such misaligned supervision leads to suboptimal performance in abstractive tasks, as the retriever fails to capture evidence passages from the corpus based on answerablility~\citep{khattab2021relevanceguided,tao2023core}. 

One straightforward solution is to annotate the answerability of passages for questions. Recent studies \citep{izacard2021distilling,sachan2021emdr2,izacard2022atlas} rely model-centric approaches to obtain strong supervision for abstractive tasks. These methods utilize iterative pipelines that leverage fine-grained supervision signals from the reader to approximately measure the answerability of retrieved passages. However, these methods require exceptionally large computational resources, which hamper their application in practical scenarios.   

Instead of pursuing such compute-intensive model-centric approaches, our work takes a step towards a data-centric approach, which aims to convert weak supervision from IR datasets into strong supervision signals for evidentiality-awareness. To this end, we present a data augmentation strategy where we augment strong distractor samples by removing evidence spans from gold evidence passages. Our strategy includes an effective approach that obtains pseudo-evidence using off-the-shelf QA model for datasets without gold annotations.
We further propose Evidentiality-Aware Dense Passage Retriever (EADPR), a novel learning approach for dense retrieval that maximally leverages augmented distractor samples to integrate evidentiality-awareness into dense passage retrievers. In EADPR, our distractor passages as both hard negatives and pseudo-positives, as the model learns to discriminate evidence passages from strong distractors (\ie, hard negatives) and distinguish between irrelevant and semantically relevant contexts (\ie, pseudo-positives). 
Using these distractors as pivots between evidence and irrelevant passages, we aim at training an effective dense retriever that ranks evidence passages higher over distractor passages.

We evaluate EADPR across multiple ODQA tasks to show that our model leads to considerable improvement in retrieval and QA performance, and that our approach can be orthogonally applied with common strategies used to train advanced retrievers such as negative sampling \citep{xiong2020ance, qu2021rocketqa}. We also conduct extensive analysis on EADPR to show that our evidentiality-aware learning shows promise for robust, efficient approach to dense passage retrieval.

\section{Preliminaries}\label{sec:related} 

A common approach to ODQA tasks usually involves utilizing external knowledge from a large corpus of texts to produce factually correct outputs~\citep{chen2020open}.
Due to the large search space in the corpus, a retriever is used in such settings to find subsets of relevant passages to questions for the expensive reader. 
The predominant approach to passage retrieval is DPR~\citep{karpukhin2020dpr}, which leverages the efficient dual-encoder architecture denoted as $[f_q, f_p]$ to encode questions and passages into a learned embedding space. 
For a question-relevant passage pair $(q_i, p^+_i)$ and a set of $N$ negative passages $p^-_j$, DPR is trained to maximize the relevance measure (\eg, the vector similarity) between the question $q_i$ and its relevant passage $p^+_i$:
\begin{multline}\label{eq:inbatch}
    \mathcal{L}(q_i, p_i^+, \{p_j^-\}_{j=1}^{N}) =  \\ 
    -\log \frac{e^{ \langle q_i,p^+_i \rangle }}{e^{ \langle q_i,p^+_i \rangle }+\sum_{j=1}^{N}e^{ \langle q_i,p^-_j \rangle }}
\end{multline}
where $ \langle q_i,p_i \rangle $ computes the relevance score between $q_i$ and $p_i$ as dot product between the question embedding $f_q(q_i)$ and the passage embedding $f_p(p_i)$~(\ie,{ $\langle q_i, p_i \rangle = f_q(q_i) \cdot f_p(p_i)$}). 

Previous studies on dense retrieval have presented some straightforward strategies to further enhance the performance of DPR. One such approach is negative sampling \citep{xiong2020ance,qu2021rocketqa}, which exploits multiple retrievers to collect informative negative samples. While earlier work uses lexical retrievers such as BM25~\citep{robertson1994bm25} for negative sampling, recent studies find that sampling hard negatives from fine-tuned encoders  \citep{humeau2020polyencoders} leads to more informative hard negative~\citep{xiong2020ance}.

Despite these efforts, it still remains a challenge to train a dense retriever to the abstractive tasks. The main obstacle arises from the lack of large-scale data with strong annotations of evidentiality ~\citep{khattab2021relevanceguided,prakash2021incomplete,tao2023core}, \ie, whether each of the passages contains evidence needed to answer the questions. To address this issue, recent studies~\citep{izacard2021distilling,sachan2021emdr2,izacard2022atlas} employ an iterative pipeline that annotates evidentiality of the passages using supervision signals from the reader. However, using these complex model-centric approaches requires a significant amount of computing resources, which obstruct their deployment in various scenarios~\citep{lindgren2021efficient,du2022glam,gao2022tevatron}. In this work, we instead study the validity of a data-centric approach to enhance the quality of IR datasets to obtain strong supervisions for passage retrieval from weak supervision in IR datasets.

\begin{figure}[t]
    
    \centering
    \includegraphics[width=\linewidth]{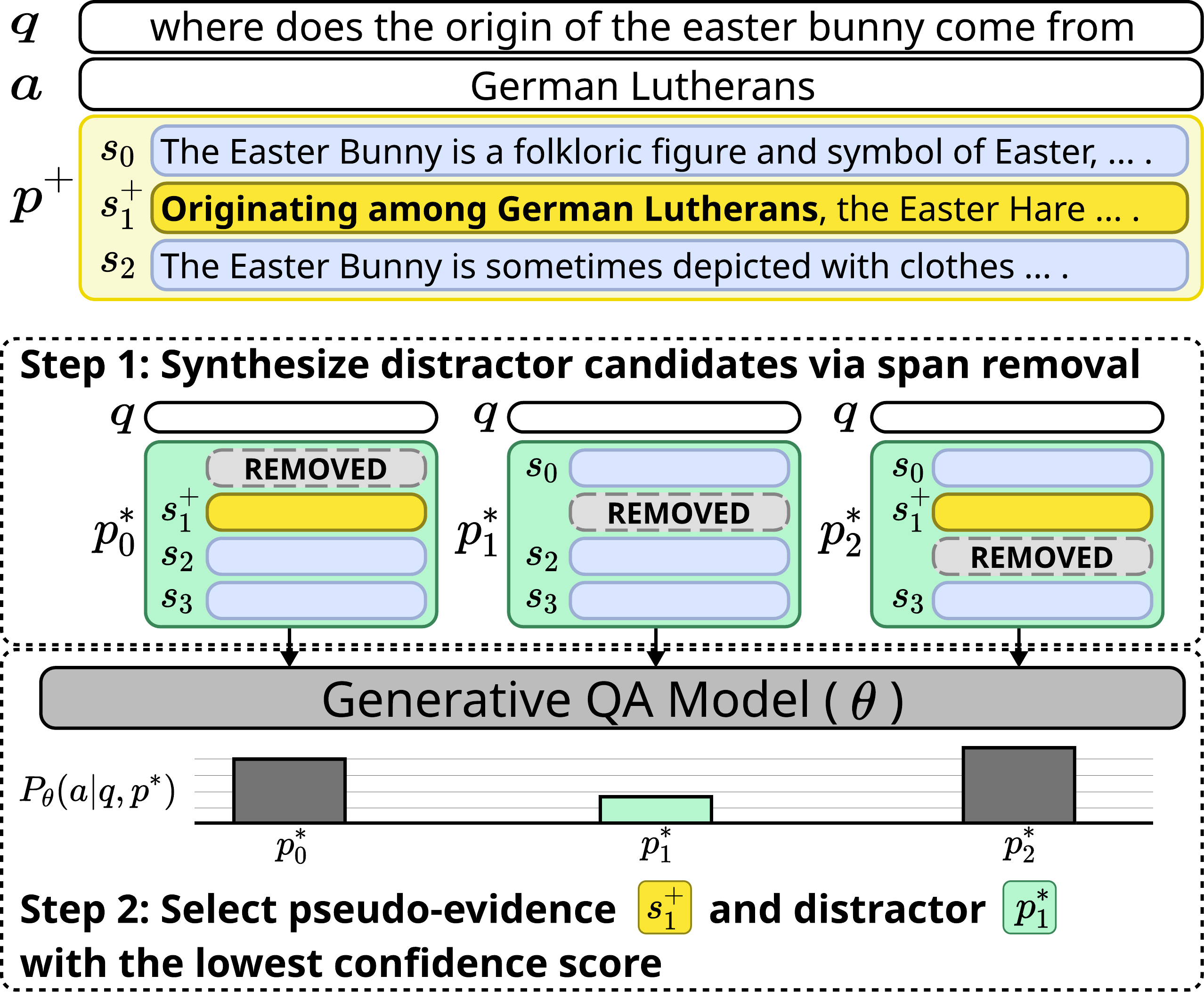}
    
    \caption{Illustration of pseudo-evidence annotation. }
    \label{fig:pseudo_evidence}
\end{figure}

\section{Methodology}\label{sec:method}

Our goal is to train a dense retriever capable of distinguishing evidence passages from distractor passages within a corpus. 
In this section, we propose Evidentiality-Aware Dense Passage Retrieval (EADPR), a novel learning approach for dense retrieval where the learned representation is conditioned on evidence spans (\ie{, \textit{positive}}) and invariant to evidentially-false contexts (\ie{, \textit{negative}}).  

\subsection{Augmenting Distractor Samples}

An intuitive approach to synthesize distractor samples is to remove evidence spans from the gold evidence passage. 
Given a question-answer passage pair $(q,p^+)$, where $p^+=[s_l;s^+;s_r]$ contains an evidence span $s^+$ to the question $q$ and evidentially-false spans $s_l$ and $s_r$, we define our \emph{distractor} sample $p^*$ as a variant of $p^+$ such that $p^*=[s_l;s_r]$.
We assume that such distractor samples are less evidential as they retain relevant semantics to the question but lack causal signals for question answering. 

One problem in distractor augmentation is that some datasets do not include annotations of evidence spans, which are costly to obtain via human annotations. To address this issue, we follow the approaches from \citet{lee2021robustifying} and incorporate pseudo-evidence annotations for distractor augmentation, as illustrated in Figure~\ref{fig:pseudo_evidence}. Specifically, we employ an off-the-shelf generative question answering (QA) model $\theta$ that takes a question and a single evidence passage as inputs to generate the answer to the input question. For a given question $q$ and its gold evidence passage $p^+$ with $n$ discrete spans, we sample $n$ distractor candidates $\{p_i^*\}_{i=1}^n$ by leaving out each of the $n$ spans from $p^+$. Each distractor candidate $p_i^*$ is then fed into the QA model with the question $q$ to compute the confidence score $P_{\theta}(a|q,p_i^*)$. We choose the candidate $p_i^*$ with the lowest confidence score as our distractor sample $p^*$, as a sharp drop in confidence score indicates that the $i$-th span is helpful in answering the question. 

In practice, we adopt UnifiedQA-T5~\citep{khasabi2020unifiedqa} as QA model and select candidates with the highest perplexity, which is commonly used as the indicator of model confidence. 

\subsection{Evidentiality-aware Learning}\label{sec:aadpr}
\begin{figure}[t]
    
    \centering
    \includegraphics[width=0.85\linewidth]{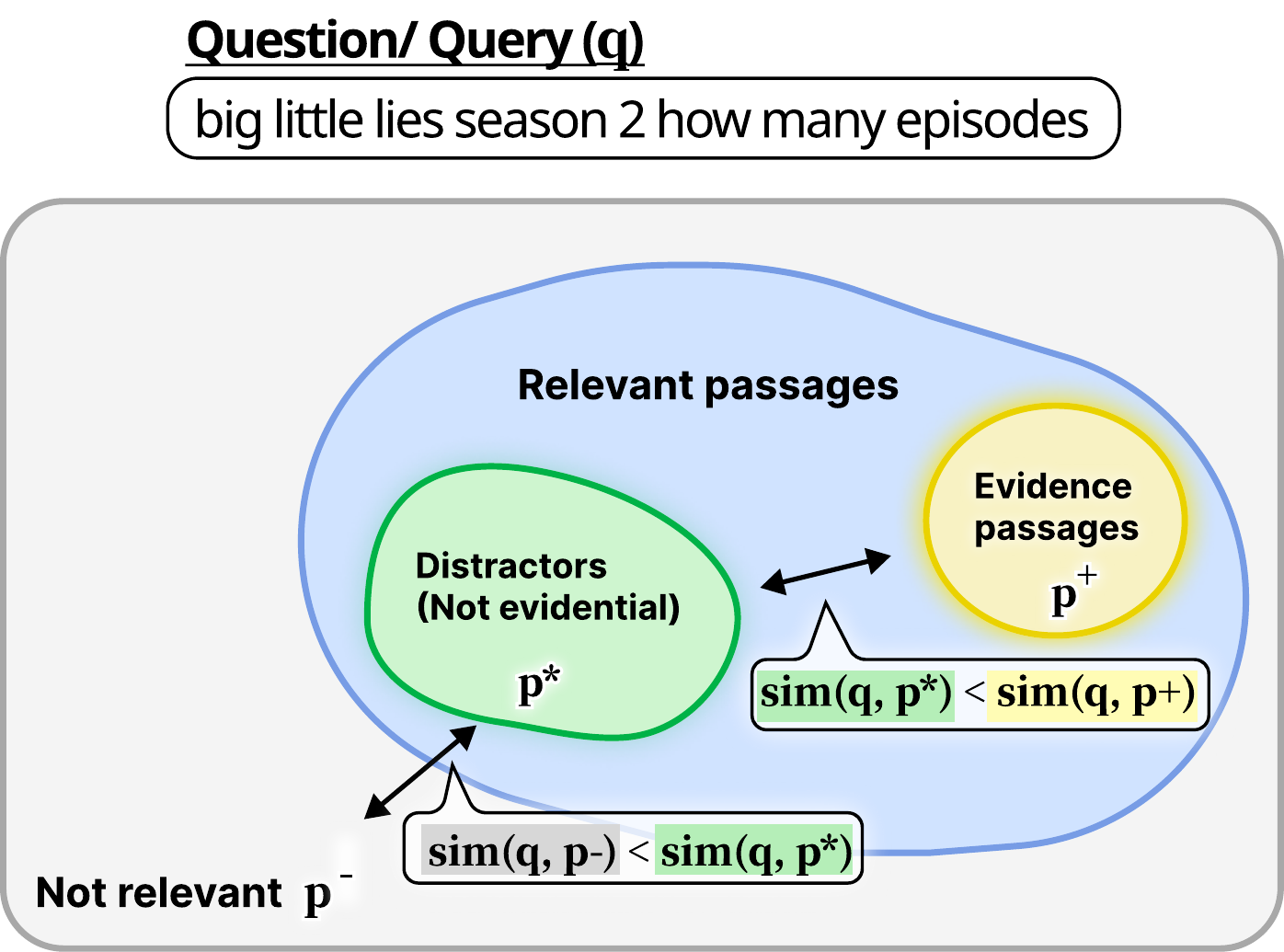}
    
    \caption{Conceptual overview of EADPR, where distractor samples serve as pivots between positive and negative passages. }
    \label{fig:cf_overview}
\end{figure}

We aim to train a retriever to learn representions of questions and passages conditioned on their evidentiality such that the retriever ranks evidence passages higher than other distractor passages. 
Our design is based on the intuition that our distractor sample, denoted as $p^*$, serves as both a hard negative and pseudo-positive, as distractor passages are still relevant to the question. 
Essentially, we model the space that is relevant but not evidential as a middle pivot point between the relevant space and the irrelevant space, as illustrated in Figure~\ref{fig:cf_overview}.

\paragraph{Distractors as Hard Negatives.}
Our distractor samples are designed to be less evidential, meaning that its content is relevant but doesn't contain the actual information for the question. As our goal is to learn a representation that reflects the evidentiality, we use these distractor samples as hard negatives. Specifically, we consider $p^*_i$ as a hard \textit{negative} sample to an anchor question $q_i$ while the original passage $p^+_i$ serves as the \textit{positive}. Thus the embedding similarity $ \langle q_i, p^*_i \rangle$ between $q_i$ and $p^*_i$ is upper bounded by $ \langle q_i, p^+_i \rangle $:
\begin{equation}\label{eq:chn_constraint}
    \langle q_i, p^+_i \rangle > \langle q_i, p^*_i \rangle
\end{equation}
Following this observation, we define \emph{Distractors-as-Hard-Negative} loss, $\mathcal{L}_{\texttt{HN}}$, to maximize the similarity between $q_i$ and $p^+_i$ while minimizing the similarity between $q_i$ and $p^*_i$. 
\begin{equation}\label{eq:chn}
    \mathcal{L}_{\texttt{HN}}(q_i, p^+_i, p^*_i)= -\log  \frac{e^{\langle q_i,p^+_i \rangle}}{e^{\langle q_i,p^+_i \rangle}+e^{\langle q_i,p^*_i \rangle}}
\end{equation}
By learning to discriminate $p^*_i$ from $p^+_i$, the model learns to minimize the mutual information between representations of questions $q_i$ and evidentially-false spans in $p^+_i$, strengthening causal effects of evidence spans in the learned embeddings.

\paragraph{Distractors as Pseudo Positives.} However, it is not sufficient to solely consider our synthetic distractors as hard negatives. Since these samples still hold relevance, our objective is to rank them lower than evidence passages but higher than irrelevant ones. While distractor samples serve as hard negatives in relation to evidence passages, they can be seen as positive samples in comparison to irrelevant ones. We refer to these samples as pseudo-positives, as semantic relevance between $q_i$ and $p^*_i$ distinguishes $p^*_i$ from other \textit{negatives} $p^-_j$, which provide noisy contexts with respect to $q_i$. Thus, the following holds for all $p^-_j$:
\begin{equation}\label{eq:cpp_constraint}
    \langle q_i, p^*_i \rangle > \langle q_i, p^-_j \rangle
\end{equation}

To incorporate this, we derive \emph{Distractors-as-Pseudo-Positives} loss, $\mathcal{L}_{\texttt{PP}}$, where the model maximizes the relative similarity between $q_i$ and $p^*_i$ with respect to negative passages $p^-_j$ and $p^*_j$ in the given batch. 
\begin{multline}\label{eq:cpp}
    \mathcal{L}_{\texttt{PP}}(q_i, p^*_i, \{p^-_j, p^*_j\}_{j \neq i}^N)=  \\  
    -\log \frac{e^{ \langle q_i,p^*_i \rangle }}{e^{ \langle q_i,p^*_i \rangle } + \sum_{j\neq i}^{N} \bigl( e^{ \langle q_i,p^-_j \rangle} + e^{ \langle q_i,p^*_j \rangle } \bigr)}
\end{multline}

Essentially, the model learns to discriminate three relevancy space check among evidential, evidentially-false, and irrelevant passages, as illustrated in Figure \ref{fig:cf_overview}.

\paragraph{Evidentiality-aware DPR.} 
From Equation~\ref{eq:chn_constraint} and \ref{eq:cpp_constraint}, we can derive that the embedding similarity $\langle q_i, p^*_i \rangle$ between questions and distractor samples are bounded by $\langle q_i, p^+_i \rangle$ and $\langle q_i, p^-_j \rangle$. Hence, they can be re-formulated as \emph{pivots} between positive and negative samples in the embedding space. 
Note that our definition of distractor samples as pivots is in line with the objective of DPR, since both inequality constraints in Equation~\ref{eq:chn_constraint} and \ref{eq:cpp_constraint} combined satisfy the below constraint in Equation~\ref{eq:inbatch}:
\begin{equation}
    \langle q_i, p^+_i \rangle > \langle q_i, p^-_j \rangle
\end{equation}

Building on top of the above idea, our training objective combines all losses from Equation~\ref{eq:chn} and \ref{eq:cpp} with the training loss in Equation~\ref{eq:dpr}.
To adapt DPR training into our setting, we further define $\mathcal{L}_{\texttt{dpr}}$ as a slight modification of DPR training objective where the distractor $p^*_i$ to the evidence passage $p^+_i$ is added as a negative:
\begin{multline}\label{eq:dpr}
    \mathcal{L}_{\texttt{dpr}}(q_i, p^+_i, p^*_i, \{p^-_j\}_{j \neq i}^N) =  \\  
    -\log{\frac{e^{\langle q_i,p^+_i\rangle}}{e^{\langle q_i,p^+_i \rangle} + \sum_{j \neq i}^{N}e^{\langle q_i,p^-_j \rangle}+\lambda e^{\langle q_i,p^*_i \rangle}}}
\end{multline}
where $\lambda < 1$ is a hyperparameter used to balance the effect from counterfactual passages as negatives in DPR training.
The final loss function $\mathcal{L}_{\texttt{eadpr}}$ is a weighted sum of all losses $\mathcal{L}_{\texttt{dpr}}$, $\mathcal{L}_{\texttt{HN}}$, and $\mathcal{L}_{\texttt{PP}}$:
\begin{equation}\label{eq:final}
    \mathcal{L}_{\texttt{eadpr}} = \mathcal{L}_{\texttt{dpr}} + \tau _1\mathcal{L}_{\texttt{HN}} + \tau _2\mathcal{L}_{\texttt{PP}}
\end{equation}
where $\tau _1, \tau _2$ are hyperparameters that determine the importance of the terms. See Appendix~\ref{sec:implementation} for details on hyperparameters.

\section{Experiments}\label{sec:exp}

\begin{table}[!t]
    \renewcommand{\arraystretch}{1.15} 
    \small
    \centering
    \begin{tabular}{lrrrr}
    \hline
    \noalign{\hrule height 0.8pt}
    \textbf{Dataset} & \textbf{Train} & \textbf{Dev} & \textbf{Test} & \textbf{Corpus}\\
    \hline
    \noalign{\hrule height 0.8pt}
    NQ   & 58,880 & 8,757 & 3,610&\multirow{3}{*}{21,015,324} \\
    TQA            & 57,369 & 8,837 & 11,313&  \\
    TREC         & 1,125 & 133   & 694& \\
    \hline
    HotpotQA            & 180,890 & 7,405 & -& 5,233,329 \\
    \hline
    \noalign{\hrule height 0.8pt}
    \end{tabular}
    \caption{Statistics of datasets used in this paper. Train, Dev, and Test represent the size of train sets, dev sets, and test sets, respectively. Corpus indicates the number of passages in the source corpus.}
    \label{tab:dataset}
\end{table}

\begin{table*}[t]\small
\renewcommand{\arraystretch}{1.2} 
\centering


\begin{tabular}{l c c c c c c c c c c}

\hline
\noalign{\hrule height 0.8pt}
  \multirow{2}{*}{\textbf{Training Strategies}}&\multirow{2}{*}{\textbf{Retriever}}&\multicolumn{3}{c}{\textbf{NQ}}&\multicolumn{3}{c}{\textbf{TQA}}&\multicolumn{3}{c}{\textbf{TREC}} \\ 
  \cmidrule{3-11}
  &&Top-1&Top-20&MRR&Top-1&Top-20&MRR&Top-1&Top-20&MRR \\ 
\hline
\noalign{\hrule height0.8pt}

\multirow{2}{*}{Vanilla Training}&DPR&31.8&74.8&43.1&38.7&74.7& 49.3&-&-&-\\
&{\our}&35.4&76.8&46.4&43.0&74.7&52.4&31.1&79.8&45.5 \\     
\hline
\multirow{2}{*}{+ BM25 Negative}&DPR&46.6&79.7&56.0&54.3&79.7&62.0&-&79.8$^\dagger$&-\\
&{\our}&48.6&80.1&57.6&\textbf{56.9}&\textbf{80.5}&\textbf{63.9}&46.8&\textbf{83.9}&\textbf{58.1} \\
\hline
\multirow{2}{*}{+ Negative Mining}&DPR&52.7&81.4&61.2&54.2&78.2&61.3&-&-&-\\
&{\our}&\textbf{54.0}&\textbf{82.6}&\textbf{62.4}&54.1&78.0&61.2&-&-&- \\

\hline
\noalign{\hrule height 0.8pt}
\end{tabular}
\caption{Passage retrieval results on single-hop QA datasets, \ie{ NQ, TriviaQA, and TREC}. Top-$k$ hit accuracy and MRR scores are reported, and the best results are marked as \textbf{bold}. $^\dagger$ indicates the performance of the baseline DPR is reported in \citep{karpukhin2020dpr}. } 

\label{table:mainResult}
\end{table*}

\begin{table*}[!ht]\small
\renewcommand{\arraystretch}{1.2} 
\centering

\begin{tabular}{l c c l l l }
\hline
\noalign{\hrule height 0.8pt}
\multirow{2}{*}{{Reader}} &\multirow{2}{*}{Training Strategies}&\multirow{2}{*}{\textbf{Retriever}}&
\multicolumn{3}{c}{\textbf{Exact Match (EM) score}} \\
\cline{4-6}
&&&Top-5 passages&Top-20 passages&Top-100 passages  \\

\midrule

\multirow{2}{*}{{DPR Reader}}&\multirow{2}{*}{Vanilla training}&DPR&{31.83}&{36.87}&{37.45}\\
&&{\our}&{34.27} (+2.44)&{{38.86} (+1.99)}&{{39.06} (+1.61)}\\
\hline

\multirow{2}{*}{{FiD$_{\text{base}}$ (T5)}}&\multirow{2}{*}{Vanilla Training}&DPR&{31.99}&{39.11}&{43.82}\\
&&{\our}&{{34.27} (+2.28) }&{{41.47} (+2.36)}&{{44.85} (+1.03)}\\
\hline

\multirow{2}{*}{{FiD$_{\text{base}}$ (T5)}}&\multirow{2}{*}{+ Negative Mining}&DPR &{38.31}&{43.13}&{45.37}\\
&&{\our}&{\textbf{40.22} (+1.91) }&{\textbf{44.32} (+1.19)}&{\textbf{47.65} (+2.28)}\\
\hline

\hline
 \noalign{\hrule height 0.8pt}
 \end{tabular}
\caption{End-to-end QA performance of retriever-reader on Natural Questions. Top-$k$ indicates the number of top retrieved passages used for reader inference. We reuse the checkpoints of DPR reader and FiD$_{\text{base}}$ (\ie{ T5-base implementation of FiD}) from \citet{karpukhin2020dpr} and \citet{izacard2021distilling}. Best scores are in \textbf{Bold}.}

\label{table:reader}
\end{table*}


\subsection{Experimental Settings}\label{sec:exp.1}
\paragraph{Dataset.} For our experiments we consider two categories of ODQA datasets, single-hop and multi-hop datasets. Single-hop datasets require the model to capture evidence that is not evidently given in the set of retrieved passages. The role of EADPR is to discriminate answer passages from distractor passages such that the answer passages are among the top-$k$ relevant contexts. Following \citet{karpukhin2020dpr}, we choose Natural Questions (NQ)~\citep{kwiatkowski2019natural}, TriviaQA (TQA)~\citep{joshi2017triviaqa}, and TREC~\citep{baudi2015curatedtrec} for evaluation and use the Wikipedia corpus of 21M passages as source passages.

On the other hand, a multi-hop QA dataset contains questions whose answers cannot be extracted from a single answer passage. We aim to assess whether the retrievers are capable of finding all evidence passages in the corpus such that the reader can derive answers by aggregating evidence from the passage set. Specifically, we evaluate our approach on HotpotQA~\citep{yang2018hotpotqa} under the \textit{full-wiki} setting, which uses the corpus of 5.2M preprocessed passages from Wikipedia for evaluation. See Appendix~\ref{appendix:dataset} for more details on datasets. Table~\ref{tab:dataset} summarizes the statistics of the datasets used in this paper.

\paragraph{Retriever Training Strategies. } 
We adopt DPR \citep{karpukhin2020dpr} as the backbone architecture for all retrievers implemented in this section. Our focus is to assess how applying EADPR affects the performance of the backbone DPR and whether EADPR is orthogonal to conventional approaches for retriever training. 

One popular data augmentation approach to enhance DPR involves negative mining~\citep{xiong2020ance, qu2021rocketqa}, which adopts additional retrievers to augment the train set with more informative negatives for retriever training. In our experiments, we first consider using BM25~\citep{robertson1994bm25} to sample hard negatives based on lexical matching. We then follow ANCE~\citep{xiong2020ance} and mine hard negatives from previous retriever checkpoints. For both cases, we mine one negative sample per query from top retrieved results of the retriever. We provide more details on our implementations in Appendix~\ref{sec:implementation}.

\subsection{Single-hop QA Benchmarks}\label{sec:exp.2}

\paragraph{Retrieval Performance. }
Table~\ref{table:mainResult} compares the performance of \our~models with the baselines on single-hop QA benchmarks. 
We observe that \our~models yield consistent performance gains over vanilla DPR under all tested conditions. 
Similar to DPR, \our~shows stronger performance when trained with hard negatives, which suggests that adding informative negative samples further boosts the discriminative power of \our. 
In the case of TriviaQA, we hypothesize that both retrievers trained on TriviaQA fail to deliver high-quality negative samples since the models are trained on the TriviaQA train set that contains false positive annotations~\citep{li2023large}.
Overall, the performance gain on \our~implies that \our~can be further improved when orthogonally applied with common training strategies for dense retrieval. 

\paragraph{End-to-End QA Performance.}

To assess the effect of EADPR on QA performance, we pair \our~into a QA system and evaluate the performance of the subsequent reader. 
Specifically, we re-use two reader models, an extractive reader from \citet{karpukhin2020dpr} and a Fusion-in-Decoder (FiD) from \citet{izacard2021fid}, and switch different retrievers to sample Top-$k$ passages for reader inference. 
We then compute Exact Match (EM) scores for the reader, which measures the proportion of questions whose answer prediction is equivalent to correct answers.
Table~\ref{table:reader} reports the QA performance of the retriever-reader pipelines. Overall, \our~consistently improves the QA performance of different readers over DPR, suggesting that EADPR benefits the subsequent readers.

\begin{table}[t]\small
\renewcommand{\arraystretch}{1.25} 
\centering

\begin{tabular}{l c c c }
 \hline
 \noalign{\hrule height 0.8pt}
 &R@2&R@10&R@20 \\ 
 \hline
 \multicolumn{3}{l}{\textit{Single-hop Retrieval}} \\ 
 ~~~DPR$^\dagger$ ~~~~~~~~ &25.2&45.4&52.1 \\
 ~~~EADPR ~~~~~~~~ &29.4&48.5&53.5 \\
 \hline
 \multicolumn{3}{l}{\textit{Multi-hop Retrieval}} \\ 
 ~~~MDR+DPR ~~~~~~~~ &47.7&61.0&65.7 \\
 ~~~MDR+EADPR ~~~~~~~~ &\textbf{58.5}&\textbf{68.1}&\textbf{71.8} \\
\hline
\noalign{\hrule height 0.8pt}
\end{tabular}
\caption{Retrieval performance of EADPR on HotpotQA. R@k indicates the proportion of questions where all annotated supporting contexts are included in top-k retrieval results. $^\dagger$ denotes the reported performance in \citet{xiong2020mdr}.} 

\label{table:multihop_retrieval}
\end{table}
\begin{table}[t]
    \renewcommand{\arraystretch}{1.5} 
    \centering
    \small
    \begin{tabular}{lccc}
    \hline
    \noalign{\hrule height 0.8pt}
    Retriever&Answer&Support&Joint \\
    \hline
    MDR+DPR & 61.0 & 61.5 & 50.2  \\
    MDR+EADPR &\textbf{ 66.1 }& \textbf{68.2} & \textbf{56.4} \\
    \hline
    \noalign{\hrule height 0.8pt}
    \end{tabular}
    \caption{Reader performance on HotpotQA dev set. We report F1 scores of the ELECTRA reader given 20 supporting contexts, which are much fewer than 100 contexts used in \citet{xiong2020mdr}.}
    \label{tab:mdr_QA}
\end{table}
\subsection{Multi-hop QA Benchmark}

We evaluate our approach on HotpotQA~\citep{yang2018hotpotqa} to assess whether EADPR better capture key evidence in multi-hop QA settings, where passages contain implicit evidence rather than answer exact match. Table~\ref{table:multihop_retrieval} compares the performance of DPR and EADPR implemented for single-hop and multi-hop retrieval. For our multi-hop retrievers, we follow \citet{xiong2020mdr} and implement multi-hop dense retriever (MDR) using EADPR. We use MDR models to produce 20 candidate contexts and feed them into an ELECTRA reader~\citep{clark2020electra}. Details on multi-hop baselines are included in Appendix~\ref{sec:implementation}.

Table~\ref{table:multihop_retrieval} shows that EADPR shows higher R@k than a vanilla DPR even without applying MDR, suggesting that our evidentiality-aware training improves the model's ability to capture key evidence without attending to exact answer match.
We also observe that incorporating EADPR into MDR leads to considerable performance gain over the standard MDR implemented using DPR, and that such gain in retrieval performance leads to improvement in QA performance, as shown in Table~\ref{tab:mdr_QA}. Full results are shown in Table~\ref{tab:mdr_QA_full}.

\section{Analysis and Discussion}\label{ref:answer_aware}

\begin{figure*}[t]

    \begin{subfigure}[b]{0.45\textwidth}
        \centering
        \includegraphics[width=\textwidth]{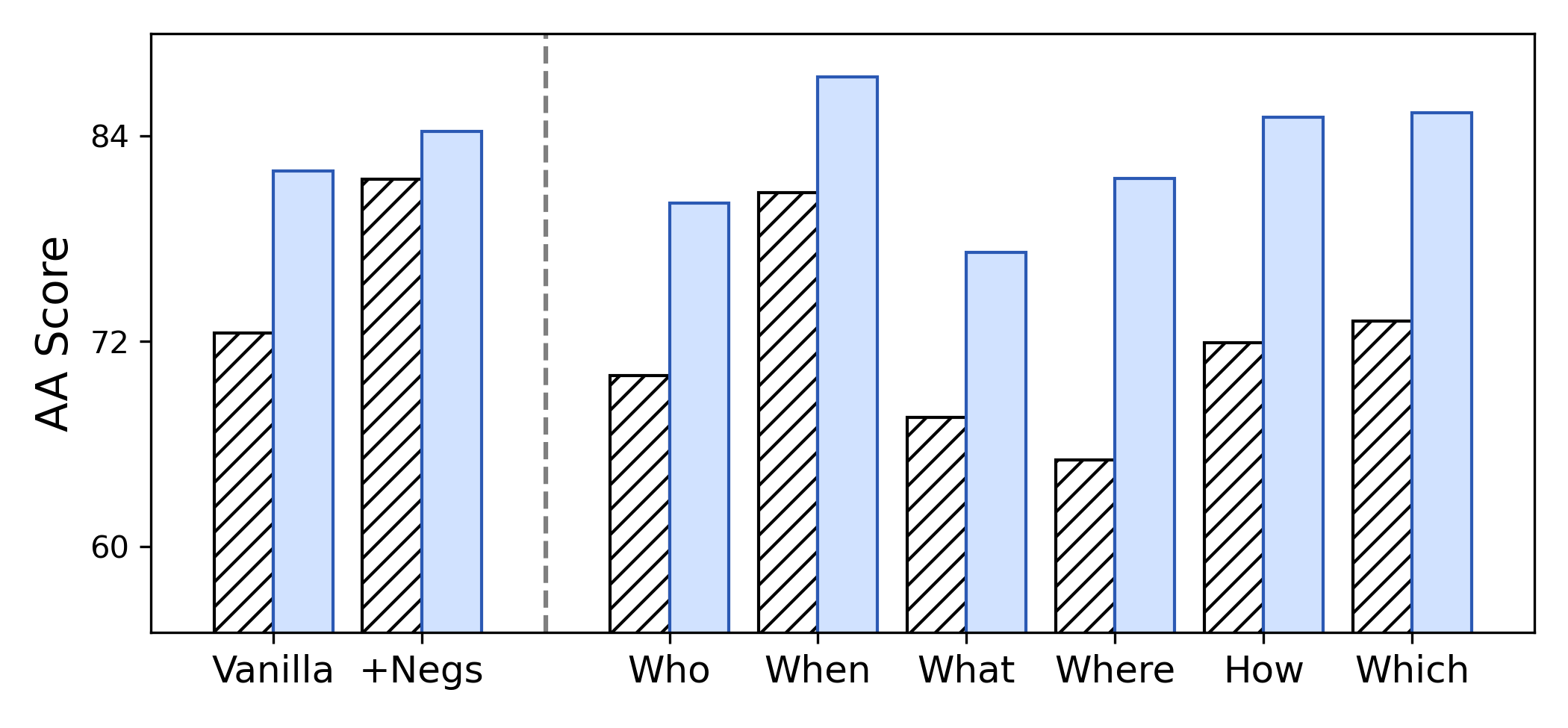}
        \caption{Answer Awareness}
        \label{fig:qtype-aa}
    \end{subfigure}
    \hfill
    \begin{subfigure}[b]{0.45\textwidth}
        \centering
        \includegraphics[width=\textwidth]{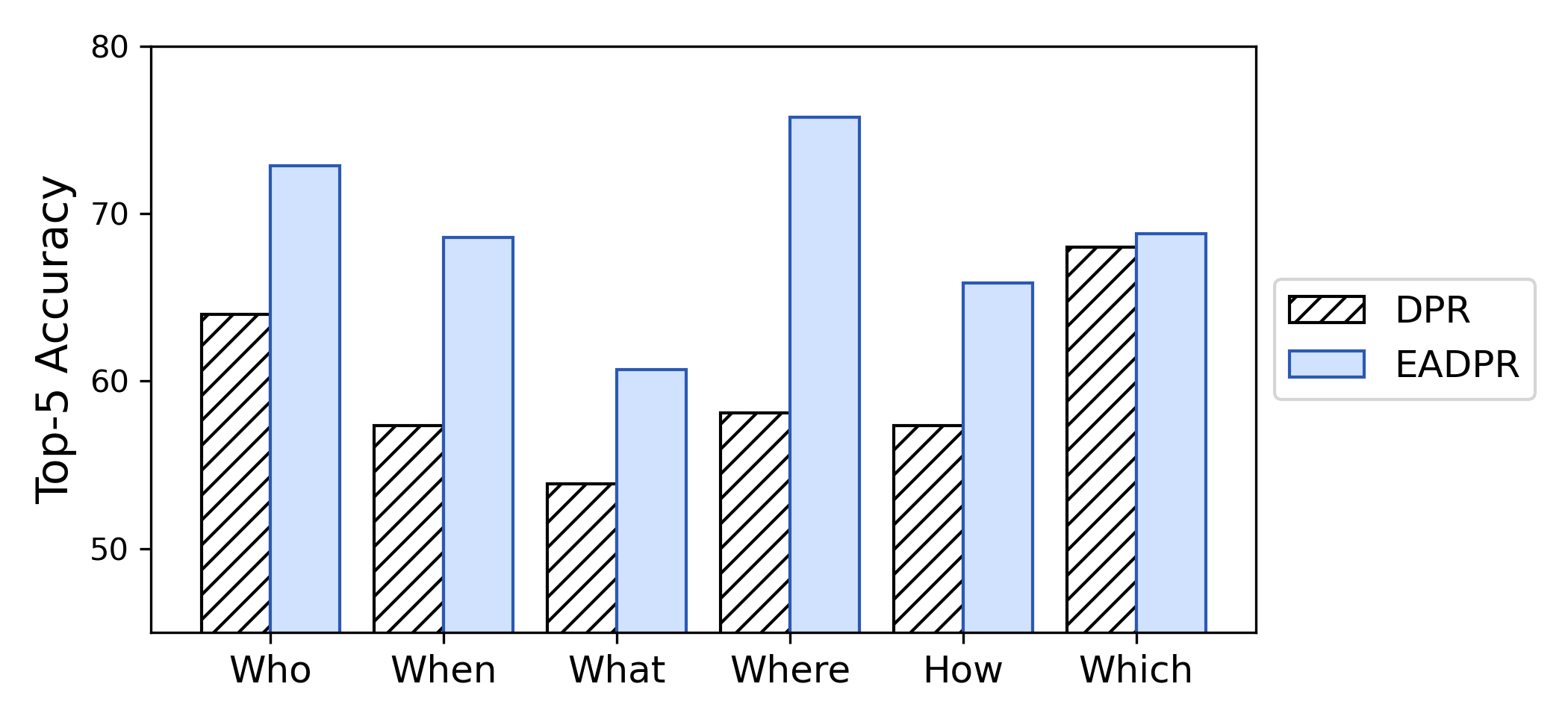}
        \caption{Retrieval Top-5 Accuracy}
        \label{fig:qtype-top5}
    \end{subfigure}
    
    \caption{AA score and retrieval top-5 hit accuracy on various question types. Note that we feed retrieved passages from DPR or \our~on the same DPR reader for inference.}
    \label{fig:qtype}
\end{figure*}

\begin{table*}[!t]\small
\renewcommand{\arraystretch}{1.5} 
\centering

\begin{tabular}{l|c}
\hline
\noalign{\hrule height 0.8pt}
    \multicolumn{2}{c}{\textbf{Question: Who died in the plane crash greys anatomy}}\\
    \hline
    \textbf{Passage Type}& \textbf{Text} \\
    \hline
    Gold passage & \multicolumn{1}{p{0.8\textwidth}}{Flight (Grey's Anatomy) ... American television medical drama Grey 's Anatomy … who are victims of an {aviation accident} fight to stay alive , {but \textbf{Dr. Lexie Grey} ( Chyler Leigh ) ultimately dies}. ... }\\
    \hline
    {DPR Top-1}&\multicolumn{1}{p{0.8\textwidth}}{Paul-Louis Halley. Socata TBM 700 {aircraft crash} on 6 December 2003, during an approach to Oxford {Airport}. The {plane} went into an uncontrolled roll, {killing Halley}, his wife, and the {pilot}. …}\\
    \hline
    DPR Top-9&\multicolumn{1}{p{0.8\textwidth}}{Flight (Grey's Anatomy).. {plane} and awakens alone in the wood; his mangled hand having been pushed through the door of the {plane}. However, none are in as bad shape as {\textbf{Lexie}, who is crushed under ...}}\\
    \hline
    \our~Top-1 &\multicolumn{1}{p{0.8\textwidth}}{Comair Flight 5191. {after the crash} to create an appropriate memorial for the victims, first responders, ... suffered serious injuries, including multiple {broken bones, a collapsed lung, and severe bleeding}. ...}\\
    \hline
    \our~Top-2 &\multicolumn{1}{p{0.8\textwidth}}{Flight (Grey's Anatomy)... {plane} and awakens alone in the wood; his mangled hand having been pushed through the door of the {plane}. However, none are in as bad shape as {\textbf{Lexie}, who is crushed under ...}}\\
    
\hline
\noalign{\hrule height 0.8pt}
\end{tabular}
\caption{An example case on `who' questions from results of DPR and \our. Answers are in \textbf{Bold}.}

\label{tab:example dpr}
\end{table*} 

\paragraph{Answer Awareness.}
To see how EADPR achieves such improvement, we conduct a fine-grained analysis to measure the model's capability of capturing evidence spans. For this purpose, we introduce an additional analytic metric, named \underline{A}nswer-\underline{A}wareness (AA) score, by measuring how frequently the model deems an answer-masked passage more relevant than its original passage. Formally, given a held-out set of $T$ pairs $(q_i,p^+_i)$ with gold answer annotations, we construct answer-masked passages $p'_i$ by removing exact answer spans from $p^+_i$. $\text{AA}$ score of a retriever is then computed as the proportion of $(q_i,p^+_i,p'_i)$ triplets where relevance scores $\langle q_i, p^+_i \rangle$ are higher than the scores $\langle q_i, p'_i \rangle$ of answer-masked passages:
\begin{equation}
    \text{AA}~\text{score} = 1 - {\sum_{i=1}^T \mathds{1}_{\langle q_i, p^+_i \rangle \leq \langle q_i,p'_i \rangle}}/ T 
\end{equation}
where $\mathds{1}_{\langle q_i, p^+_i \rangle \leq \langle q_i,p'_i \rangle}$ is an indicator if $\langle q_i, p^+_i \rangle$ is smaller than $\langle q_i,p'_i \rangle$. 
To measure $\text{AA}$ score, we reuse the 1,382 gold $(q, p^+, p')$ triplets from NQ test set used in Section~\ref{sec:preliminary}.

Figure~\ref{fig:qtype} compares the $\text{AA}$ score of EADPR with DPR trained under the same conditions (\ie, training strategies).  
We first observe that AA scores of a vanilla DPR significantly fall behind the theoretical upper bound, which indicates that the relevance measurement learned in DPR may not effectively capture the evidentiality-awareness such that retrievers constantly rank positive passages higher than counterfactual passages. While common training strategies such as negative sampling lead to some increase in $\text{AA}$ scores, there is still substantial room for improvement towards building an evidentiality-aware retriever. One the other hand, EADPR brings further gain in $\text{AA}$ scores, showing that our data-centric approach is effective in enhancing evidentiality-awareness.

In Figure~\ref{fig:qtype},  we further break down the the gold $(q,p^+,p')$ triplets with respect to their question types and measure AA scores of DPR and EADPR on subsets of test samples of different question types, \ie, who, when, what, where, how, and which. Overall, we see that $\text{AA}$ scores of DPR vary significantly across different question types, ranging from 65.07\% to 80.68\%. On the other hand, EADPR achieves significant improvements in $\text{AA}$ scores for all question types and consistently shows better retrieval performance.

Among all question types, we see that DPR shows particularly low AA scores on who-, what-, and where-questions, whose answers tend to refer to named entities, \ie, names of people, locations, and objects. Our hypothesis is that DPR often fails to identify the presence of target entities, which serve as causal features in evidence passages. Table~\ref{tab:example dpr} shows an example of the retrieval results, illustrating the problem of named entities for DPR. While DPR is capable of retrieving passages with relevant semantics such as aircraft crash, it fails to identify key named entities in the question such as Greys Anatomy. In contrast, we observe EADPR ranks evidence passages with key entities higher than DPR (\ie, Top-2 from EADPR compared to Top-9 from DPR), suggesting that EADPR learns to differentiate evidence passages from their distractors in which key entities are absent. 
In some sense, our approach is in line with previous methods based on salient span masking~\citep{guu2020retrieval,sachan2021end}, where the retriever is trained to predict masked salient spans with the help of a reader.

\begin{figure}[t]
    \centering
    \includegraphics[width=\linewidth]{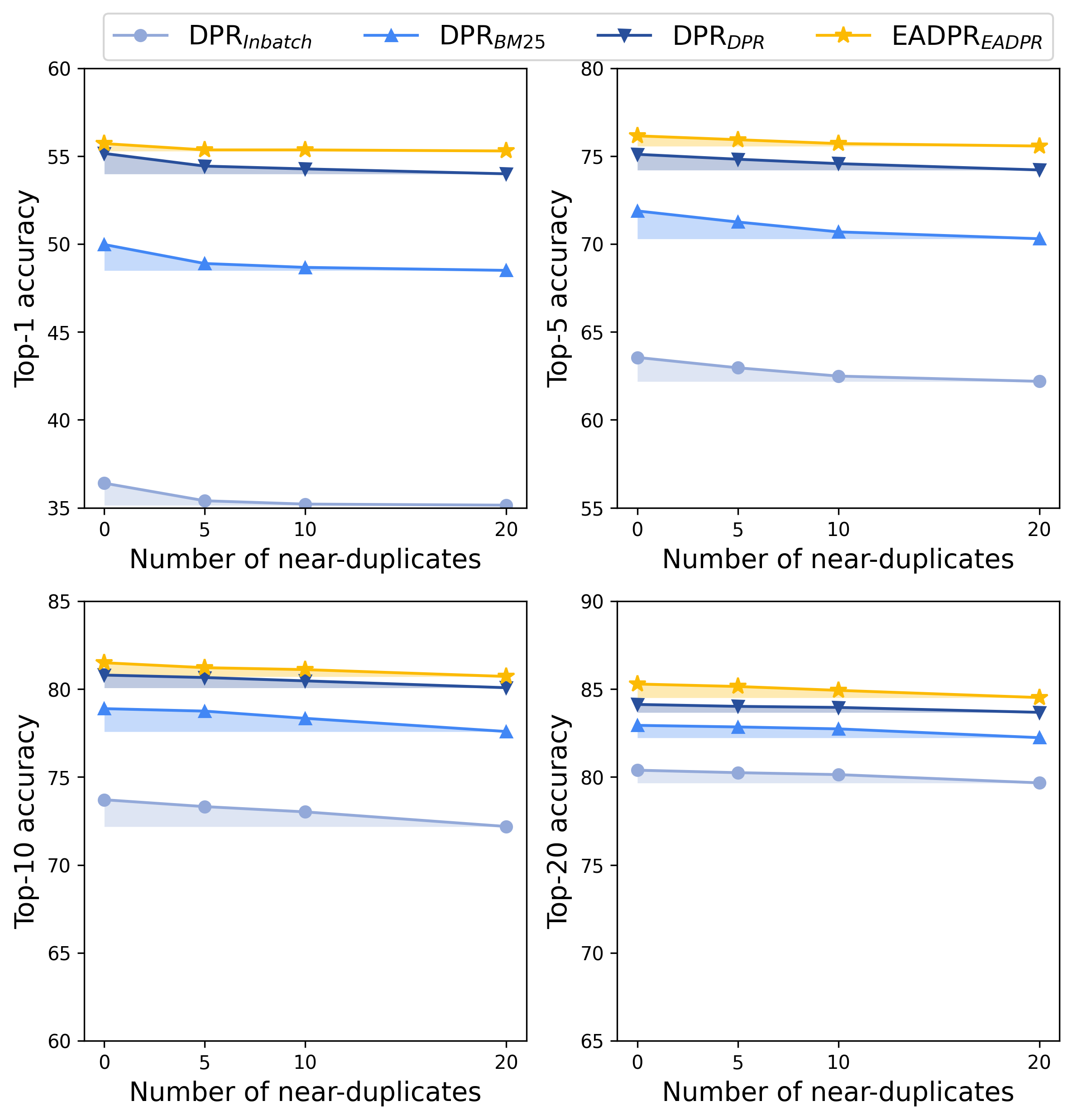}
    \caption{ Retrieval accuracy at Top-$k$ with varying number of near-duplicates on Natural Questions dataset. 
    Colored area illustrates the degree of the performance drop. 
    }
    \label{figure:near-duplicates}
\end{figure}

\paragraph{Robustness.}\label{sec:preliminary}
We have assumed that EADPR learns to discriminate between evidence and distractor passages. To validate this assumption, we perform a simulation test in which we synthesize and add distractor passages into the corpus to measure the robustness of EADPR to these samples. This scenario is often encountered in real-world corpora, where there is a surplus of passages with similar contexts but lack definitive evidence ~\citep{spirin2012spam,pan2023risk,goldstein2023generative}.

Specifically, we create plausible distractor passages using a large language model (\ie, ChatGPT). The model is prompted to generate \emph{near-duplicate} samples for each query that mimic the context of evidence passage but leave out the key evidence. We collect these near-duplicates for 1,382 test queries from NQ with annotated evidence passages and include at most 20 near-duplicates per query.

Figure~\ref{figure:near-duplicates} shows the performance of the dense retrievers on text corpora with varying number of near-duplicates. We observe that the retrieval performance (\ie, Top-$k$ accuracy) decreases substantially when given more near-duplicates, indicating 
that dense retrievers are vulnerable to the presence of distractor passages. 
On the other hand, we observe that EADPR is relatively robust against the effect from additional distractor samples, showing promise for robust passage retrieval on a noisy real-world corpus.

\paragraph{Resource and Label Efficiency.}

\begin{figure}[t]
    \begin{subfigure}[b]{0.22\textwidth}
        \centering
        \includegraphics[width=\textwidth]{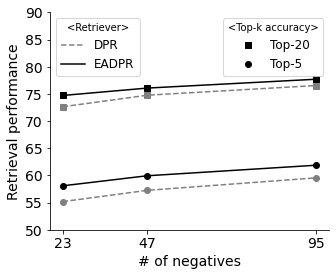}
        \caption{Number of negatives}
        \label{fig:label_negative}
    \end{subfigure}
    \hfill
    \begin{subfigure}[b]{0.22\textwidth}
        \centering
            \includegraphics[width=\textwidth]{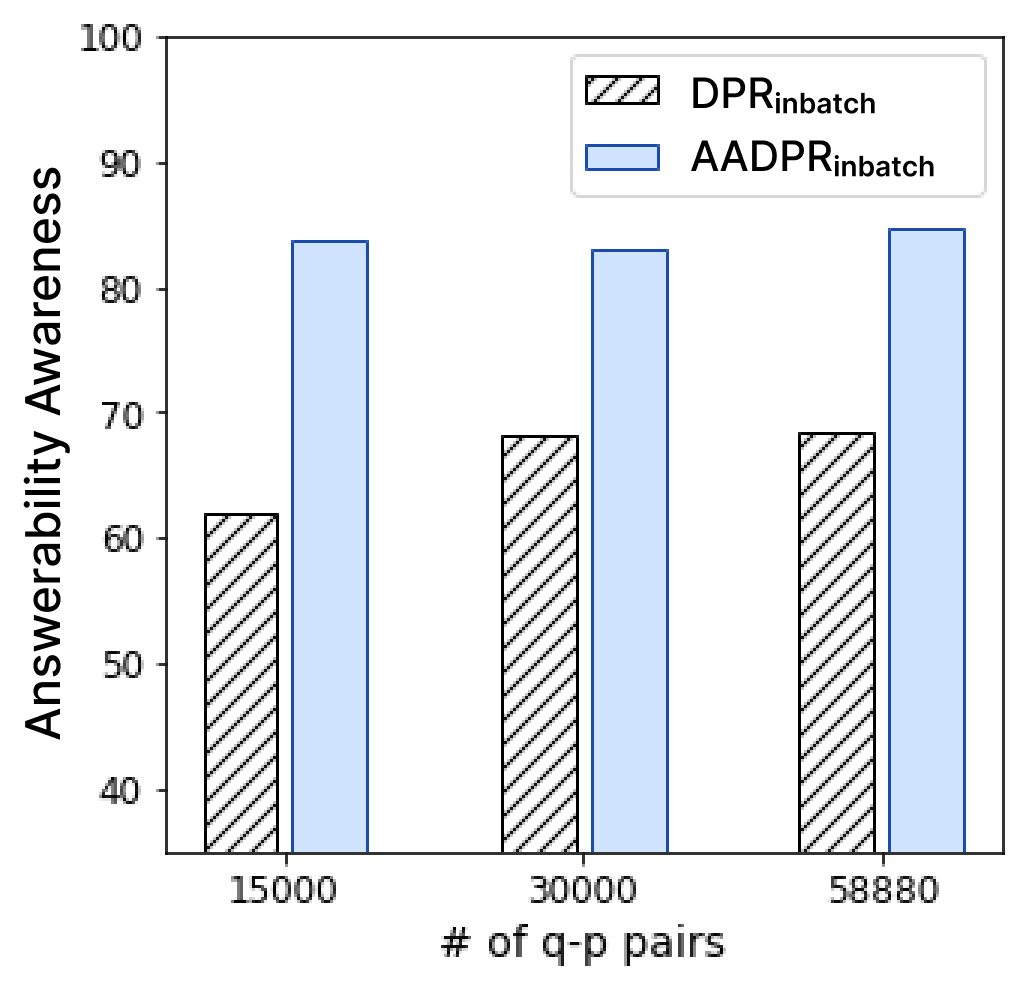}
            \caption{Training samples}
            \label{fig:label_positive}
    \end{subfigure}
    \caption{Performance of DPR and EADPR on varying numbers of (a) negatives and (b) training samples.}
    \label{fig:analysis_section}
\end{figure}

We posit that the benefit of \our~lies in the label efficiency, as counterfactual samples serve as both hard negatives and pseudo-positives in \our. To validate this assumption, we train EADPR with fewer (a) negative samples and (b) training instances.
Figure~\ref{fig:label_negative} shows that EADPR trained with fewer negatives (\eg, 23) yields performance comparable to a vanilla DPR trained with more negatives (\eg, 47). 
Meanwhile, we see in Figure~\ref{fig:label_positive} that EADPR consistently shows higher AA score over DPR when using fewer training samples (\eg, 15k and 30k). These findings support our assumption on the efficiency of EADPR.

\begin{table}[t]\small
\centering
\renewcommand{\arraystretch}{1.35} 
\begin{tabular}{l c c c c}
\hline
    \noalign{\hrule height 0.8pt}
    \multirow{2}{*}{\textbf{Retriever}}&\multicolumn{4}{c}{\textbf{Natural Questions}}\\
    \cline{2-5}
    &Top-1 &Top-5 &Top-20 &Top-100\\
\hline
\noalign{\hrule height0.8pt}
     $\mathcal{L}_{\texttt{dpr}}$ 
     &31.77&58.12&74.76&84.07 \\
\hline
     ~ $+ \mathcal{L}_{\texttt{PP}}$ 
     &32.08&58.64&75.32&83.82 \\
     ~ $+ \mathcal{L}_{\texttt{HN}}$ 
     &31.85&59.53&75.57&84.43 \\
     ~ \textbf{$+ \mathcal{L}_{\texttt{PP}} + \mathcal{L}_{\texttt{HN}}$}
     &35.35&61.55 &76.81&85.87 \\
\hline
\noalign{\hrule height 0.8pt}
\end{tabular}
\caption{Ablation studies on the training objective.}

\label{tables:ablation_studies_loss}
\end{table}

\paragraph{Effect of Counterfactual Samples as Pivots.}
We conduct an ablation studies on the learning objective in Equation~\ref{eq:final} to study the effect of using counterfactual samples as pivots (\ie, both pseudo-positives and hard negatives) on DPR training. Specifically, we consider the following modifications to the objective function, 1) $\mathcal{L}_{\texttt{dpr}} + \mathcal{L}_{\texttt{PP}}$, and 2) $\mathcal{L}_{\texttt{dpr}} + \mathcal{L}_{\texttt{HN}}$. Table~\ref{tables:ablation_studies_loss} compares all baselines with \our~and DPR. We find that all modifications do not bring much improvement to DPR without either 
$\mathcal{L}_{\texttt{PP}}$ or $\mathcal{L}_{\texttt{HN}}$. In contrast, EADPR consistently outperforms DPR and all its variants, suggesting that using counterfactual samples as pivots is crucial in \our. 

\section{Related Work}

\paragraph{Dense Retrieval. } Dense retrieval aims at retrieving information based on semantic matching by mapping questions and contexts into a learned embedding space~\cite{karpukhin2020dpr,lee2019ict}. Earlier attempts to enhance dense retrievers have drawn inspiration from studies on learning to rank \citep{liu2009learningtorank}, improving the performance of dual encoders via methods such as negative sampling (\eg, ANCE~\citep{xiong2020ance} and RocketQA~\citep{qu2021rocketqa}). More recent approaches are founded upon knowledge distillation \citep{hinton2015distilling}, which constructs an iterative pipeline of retrievers and readers such that the retriever learn from the reader's predictions on the evidentiality of passages (\eg, cross-attention in \citet{izacard2021distilling}, model confidence in ATLAS~\citep{izacard2022atlas} and REPLUG~\citep{shi2023replug}).

\paragraph{Counterfactual learning in NLP.} Counterfactual learning has been a useful tool in enhancing the robustness and fairness in representation learning by attending to causal features \citep{johansson2016counterfactual,feder2022causal}. These studies define counterfactual intervention based on causal features and train models using counterfactual samples, which are minimally dissimlar but lead to different (\ie, counterfactual) outcome \citep{chen2020counterfactual,choi2020less,choi2022c2l}. 
By learning from counterfactual samples, these approaches aim to build models that rely more on causal relationship between observations and labels.
Our work stems from this line of research, as we introduce assumptions on causal signals in passage retrieval for knowledge-intensive task. 

\section{Conclusion}
In this work, we address the misalignment problem in dense retrievers for abstractive QA tasks, where relevance supervisions from IR datasets are not well-aligned with answerability of passages for questions.
To overcome the abstractiveness of ODQA tasks, we present \our, which augments distractor samples to train an evidentiality-aware retriever by learning to distinguish between evidence and distractor samples.  Our experiments show promising results in many ODQA tasks, indicating that EADPR not only enhances model performance on both retrieval and downstream tasks but also improves robustness to distractors.

\section*{Limitations}
Below we summarize some limitations of our work and discuss potential directions to improve it: 
(i) Our definition of causal signals in answerable passages has been limited to answer sentences that contain exact matches of gold answers. While simple and efficient, our counterfactual sampling strategy leaves room for improvement, and more elaborate construction methods would lead to better counterfactual samples and further enhance the performance of \our. 
(ii) We observe that AA scores in Section~\ref{ref:answer_aware} are not well calibrated with the downstream performance of the retriever, which limits the practical usefulness of AA score as an indicator of the model performance. In future work, we aim to refine the definition of AA score such that it serves as a formal evaluation metrics for dense retrieval. 

\section*{Broader Impact and Ethics Statement}
Our work re-examines the evidentiality-awareness of the dense retrievers and seeks to mitigate undesired model biases to false positives, or contexts in candidate passages with no evidence. While we have focused solely on the effectiveness of our approach on ODQA, we believe that the concept of distractor samples as pivots can be further explored in other representation learning tasks such as response retrieval for dialogue systems.

Meanwhile, our work shares the typical risks towards misinformation from common dense retrieval models~\citep{qu2021rocketqa, santhanam2022colbertv2} as our implementation follows the common design based on dual encoders. Our work takes a step towards minimizing such risks from the retriever, but we note that there is still much work needed from the community to ensure the faithfulness of dense retrievers, particularly in specialized domains with insufficient data. 

\section*{Acknowledgement}
This work was supported by Institute of Information \& Communications Technology Planning \& Evaluation (IITP) grant funded by the Korean government (MSIT) (No.2020-0-01361, Artificial Intelligence Graduate School Program (Yonsei University)) and (No.2021-0-02068, Artificial Intelligence Innovation Hub) and (No.2022-0-00077, AI Technology Development for Commonsense Extraction, Reasoning, and Inference from Heterogeneous Data). Jinyoung Yeo is a corresponding author.

\bibliography{anthology,custom}

\begin{thebibliography}{48}
\expandafter\ifx\csname natexlab\endcsname\relax\def\natexlab#1{#1}\fi

\bibitem[{Asai et~al.(2020)Asai, Hashimoto, Hajishirzi, Socher, and
  Xiong}]{asai2020learning}
Akari Asai, Kazuma Hashimoto, Hannaneh Hajishirzi, Richard Socher, and Caiming
  Xiong. 2020.
\newblock Learning to retrieve reasoning paths over wikipedia graph for
  question answering.
\newblock In \emph{Proceedings of ICLR}.

\bibitem[{Bajaj et~al.(2016)Bajaj, Campos, Craswell, Deng, Gao, Liu, Majumder,
  McNamara, Mitra, Nguyen, Rosenberg, Song, Stoica, Tiwary, and
  Wang}]{nguyen2016ms}
Payal Bajaj, Daniel Campos, Nick Craswell, Li~Deng, Jianfeng Gao, Xiaodong Liu,
  Rangan Majumder, Andrew McNamara, Bhaskar Mitra, Tri Nguyen, Mir Rosenberg,
  Xia Song, Alina Stoica, Saurabh Tiwary, and Tong Wang. 2016.
\newblock Ms marco: A human generated machine reading comprehension dataset.
\newblock \emph{arXiV preprint}.

\bibitem[{Baudi{\v{s}} and {\v{S}}ediv{\'y}(2015)}]{baudi2015curatedtrec}
Petr Baudi{\v{s}} and Jan {\v{S}}ediv{\'y}. 2015.
\newblock Modeling of the question answering task in the yodaqa system.
\newblock In \emph{Experimental IR Meets Multilinguality, Multimodality, and
  Interaction}.

\bibitem[{Bird et~al.(2009)Bird, Klein, and Loper}]{bird2009natural}
Steven Bird, Ewan Klein, and Edward Loper. 2009.
\newblock \emph{Natural language processing with Python: analyzing text with
  the natural language toolkit}.
\newblock O'Reilly Media, Inc.

\bibitem[{Chen et~al.(2017)Chen, Fisch, Weston, and Bordes}]{chen2017reading}
Danqi Chen, Adam Fisch, Jason Weston, and Antoine Bordes. 2017.
\newblock Reading wikipedia to answer open-domain questions.
\newblock In \emph{Proceedings of ACL}.

\bibitem[{Chen and Yih(2020)}]{chen2020open}
Danqi Chen and Wen-tau Yih. 2020.
\newblock Open-domain question answering.
\newblock In \emph{Proceedings of ACL}.

\bibitem[{Chen et~al.(2020)Chen, Yan, Xiao, Zhang, Pu, and
  Zhuang}]{chen2020counterfactual}
Long Chen, Xin Yan, Jun Xiao, Hanwang Zhang, Shiliang Pu, and Yueting Zhuang.
  2020.
\newblock Counterfactual samples synthesizing for robust visual question
  answering.
\newblock In \emph{Proceedings of CVPR}.

\bibitem[{Choi et~al.(2022)Choi, Jeong, Han, and Hwang}]{choi2022c2l}
Seungtaek Choi, Myeongho Jeong, Hojae Han, and Seung-won Hwang. 2022.
\newblock C2l: Causally contrastive learning for robust text classification.
\newblock In \emph{Proceedings of AAAI}.

\bibitem[{Choi et~al.(2020)Choi, Park, Yeo, and Hwang}]{choi2020less}
Seungtaek Choi, Haeju Park, Jinyoung Yeo, and Seung-won Hwang. 2020.
\newblock Less is more: Attention supervision with counterfactuals for text
  classification.
\newblock In \emph{Proceedings of EMNLP}.

\bibitem[{Clark et~al.(2020)Clark, Luong, Le, and Manning}]{clark2020electra}
Kevin Clark, Minh-Thang Luong, Quoc~V. Le, and Christopher~D. Manning. 2020.
\newblock \href {https://openreview.net/pdf?id=r1xMH1BtvB} {{ELECTRA}:
  Pre-training text encoders as discriminators rather than generators}.
\newblock In \emph{Proceedings of ICLR}.

\bibitem[{Devlin et~al.(2019)Devlin, Chang, Lee, and
  Toutanova}]{devlin2019bert}
Jacob Devlin, Ming-Wei Chang, Kenton Lee, and Kristina Toutanova. 2019.
\newblock \href {https://doi.org/10.18653/v1/N19-1423} {{BERT}: Pre-training of
  deep bidirectional transformers for language understanding}.
\newblock In \emph{Proceedings of NAACL}.

\bibitem[{Du et~al.(2022)Du, Huang, Dai, Tong, Lepikhin, Xu, Krikun, Zhou, Yu,
  Firat et~al.}]{du2022glam}
Nan Du, Yanping Huang, Andrew~M Dai, Simon Tong, Dmitry Lepikhin, Yuanzhong Xu,
  Maxim Krikun, Yanqi Zhou, Adams~Wei Yu, Orhan Firat, et~al. 2022.
\newblock Glam: Efficient scaling of language models with mixture-of-experts.
\newblock In \emph{Proceedings of ICML}.

\bibitem[{Feder et~al.(2022)Feder, Keith, Manzoor, Pryzant, Sridhar,
  Wood-Doughty, Eisenstein, Grimmer, Reichart, Roberts, Stewart, Veitch, and
  Yang}]{feder2022causal}
Amir Feder, Katherine~A. Keith, Emaad Manzoor, Reid Pryzant, Dhanya Sridhar,
  Zach Wood-Doughty, Jacob Eisenstein, Justin Grimmer, Roi Reichart,
  Margaret~E. Roberts, Brandon~M. Stewart, Victor Veitch, and Diyi Yang. 2022.
\newblock Causal inference in natural language processing: Estimation,
  prediction, interpretation and beyond.
\newblock \emph{Transactions of the Association for Computational Linguistics}.

\bibitem[{Gao et~al.(2022)Gao, Ma, Lin, and Callan}]{gao2022tevatron}
Luyu Gao, Xueguang Ma, Jimmy Lin, and Jamie Callan. 2022.
\newblock Tevatron: An efficient and flexible toolkit for dense retrieval.
\newblock \emph{arXiV preprint}.

\bibitem[{Goldstein et~al.(2023)Goldstein, Sastry, Musser, DiResta, Gentzel,
  and Sedova}]{goldstein2023generative}
Josh~A. Goldstein, Girish Sastry, Micah Musser, Renee DiResta, Matthew Gentzel,
  and Katerina Sedova. 2023.
\newblock Generative language models and automated influence operations:
  Emerging threats and potential mitigations.
\newblock \emph{arXiV preprint}.

\bibitem[{Guu et~al.(2020)Guu, Lee, Tung, Pasupat, and
  Chang}]{guu2020retrieval}
Kelvin Guu, Kenton Lee, Zora Tung, Panupong Pasupat, and Mingwei Chang. 2020.
\newblock Retrieval augmented language model pre-training.
\newblock In \emph{Proceedings of ICML}.

\bibitem[{Hinton et~al.(2015)Hinton, Vinyals, and Dean}]{hinton2015distilling}
Geoffrey Hinton, Oriol Vinyals, and Jeff Dean. 2015.
\newblock Distilling the knowledge in a neural network.
\newblock In \emph{NIPS Deep Learning and Representation Learning Workshop}.

\bibitem[{Humeau et~al.(2020)Humeau, Shuster, Lachaux, and
  Weston}]{humeau2020polyencoders}
Samuel Humeau, Kurt Shuster, Marie-Anne Lachaux, and Jason Weston. 2020.
\newblock Poly-encoders: Transformer architectures and pre-training strategies
  for fast and accurate multi-sentence scoring.
\newblock In \emph{Proceedings of ICLR}.

\bibitem[{Izacard and Grave(2021{\natexlab{a}})}]{izacard2021distilling}
Gautier Izacard and Edouard Grave. 2021{\natexlab{a}}.
\newblock Distilling knowledge from reader to retriever for question answering.
\newblock In \emph{Proceedings of ICLR}.

\bibitem[{Izacard and Grave(2021{\natexlab{b}})}]{izacard2021fid}
Gautier Izacard and Edouard Grave. 2021{\natexlab{b}}.
\newblock Leveraging passage retrieval with generative models for open domain
  question answering.
\newblock In \emph{Proceedings of EACL}.

\bibitem[{Izacard et~al.(2022)Izacard, Lewis, Lomeli, Hosseini, Petroni,
  Schick, Dwivedi-Yu, Joulin, Riedel, and Grave}]{izacard2022atlas}
Gautier Izacard, Patrick Lewis, Maria Lomeli, Lucas Hosseini, Fabio Petroni,
  Timo Schick, Jane Dwivedi-Yu, Armand Joulin, Sebastian Riedel, and Edouard
  Grave. 2022.
\newblock \href {http://arxiv.org/abs/2208.03299} {Few-shot {Learning} with
  {Retrieval} {Augmented} {Language} {Models}}.
\newblock \emph{arXiv preprint}.

\bibitem[{Johansson et~al.(2016)Johansson, Shalit, and
  Sontag}]{johansson2016counterfactual}
Fredrik Johansson, Uri Shalit, and David Sontag. 2016.
\newblock Learning representations for counterfactual inference.
\newblock In \emph{Proceedings of ICML}.

\bibitem[{Johnson and Douze(2019)}]{johnson2019billion}
Jeff Johnson and Herv{\'e} Douze, Matthijs~andJ{\'e}gou. 2019.
\newblock Billion-scale similarity search with {GPUs}.
\newblock \emph{IEEE Transactions on Big Data}, 7(3):535--547.

\bibitem[{Joshi et~al.(2017)Joshi, Choi, Weld, and
  Zettlemoyer}]{joshi2017triviaqa}
Mandar Joshi, Eunsol Choi, Daniel Weld, and Luke Zettlemoyer. 2017.
\newblock {T}rivia{QA}: A large scale distantly supervised challenge dataset
  for reading comprehension.
\newblock In \emph{Proceedings of ACL}.

\bibitem[{Karpukhin et~al.(2020)Karpukhin, Oguz, Min, Lewis, Wu, Edunov, Chen,
  and Yih}]{karpukhin2020dpr}
Vladimir Karpukhin, Barlas Oguz, Sewon Min, Patrick Lewis, Ledell Wu, Sergey
  Edunov, Danqi Chen, and Wen-tau Yih. 2020.
\newblock Dense passage retrieval for open-domain question answering.
\newblock In \emph{Proceedings of EMNLP}.

\bibitem[{Khashabi et~al.(2020)Khashabi, Min, Khot, Sabhwaral, Tafjord, Clark,
  and Hajishirzi}]{khasabi2020unifiedqa}
D.~Khashabi, S.~Min, T.~Khot, A.~Sabhwaral, O.~Tafjord, P.~Clark, and
  H.~Hajishirzi. 2020.
\newblock Unifiedqa: Crossing format boundaries with a single qa system.
\newblock In \emph{Findings of EMNLP}.

\bibitem[{Khattab et~al.(2020)Khattab, Potts, and
  Zaharia}]{khattab2021relevanceguided}
Omar Khattab, Christopher Potts, and Matei Zaharia. 2020.
\newblock Relevance-guided supervision for openqa with colbert.
\newblock \emph{Transactions of the Association for Computational Linguistics}.

\bibitem[{Kwiatkowski et~al.(2019)Kwiatkowski, Palomaki, Redfield, Collins,
  Parikh, Alberti, Polosukhin, Devlin, Lee, Toutanova, Jones, Kelcey, Dai,
  Uszkoreit, Le, and Petrov}]{kwiatkowski2019natural}
Tom Kwiatkowski, Jennimaria Palomaki, Olivia Redfield, Michael Collins, Ankur
  Parikh, Danielle Alberti, Chris~andEpstein, Illia Polosukhin, Jacob Devlin,
  Kenton Lee, Kristina Toutanova, Llion Jones, Ming-Wei Kelcey,
  Matthewand~Chang, Andrew~M. Dai, Jakob Uszkoreit, Quoc Le, and Slav Petrov.
  2019.
\newblock Natural questions: A benchmark for question answering research.
\newblock \emph{Transactions of the Association for Computational Linguistics},
  pages 452--466.

\bibitem[{Lee et~al.(2019)Lee, Chang, and Toutanova}]{lee2019ict}
Kenton Lee, Ming-Wei Chang, and Kristina Toutanova. 2019.
\newblock Latent retrieval for weakly supervised open domain question
  answering.
\newblock In \emph{Proceedings of ACL}.

\bibitem[{Lee et~al.(2021)Lee, Hwang, Han, and Lee}]{lee2021robustifying}
Kyungjae Lee, Seung-won Hwang, Sang-eun Han, and Dohyeon Lee. 2021.
\newblock Robustifying multi-hop {QA} through pseudo-evidentiality training.
\newblock In \emph{Proceedings of ACL}.

\bibitem[{Li et~al.(2023)Li, Rawat, Zaheer, Wang, Lukasik, Veit, Yu, and
  Kumar}]{li2023large}
Daliang Li, Ankit~Singh Rawat, Manzil Zaheer, Xin Wang, Michal Lukasik, Andreas
  Veit, Felix Yu, and Sanjiv Kumar. 2023.
\newblock Large language models with controllable working memory.
\newblock In \emph{Findings of ACL}.

\bibitem[{Lindgren et~al.(2021)Lindgren, Reddi, Guo, and
  Kumar}]{lindgren2021efficient}
Erik Lindgren, Sashank Reddi, Ruiqi Guo, and Sanjiv Kumar. 2021.
\newblock Efficient training of retrieval models using negative cache.
\newblock \emph{Advances in Neural Information Processing Systems},
  34:4134--4146.

\bibitem[{Liu(2009)}]{liu2009learningtorank}
Tie-Yan Liu. 2009.
\newblock \href {https://doi.org/10.1561/1500000016} {Learning to rank for
  information retrieval}.
\newblock \emph{Foundations and Trends in Information Retrieval},
  3(3):225–331.

\bibitem[{Pan et~al.(2023)Pan, Pan, Chen, Nakov, Kan, and Wang}]{pan2023risk}
Yikang Pan, Liangming Pan, Wenhu Chen, Preslav Nakov, Min-Yen Kan, and
  William~Yang Wang. 2023.
\newblock \href {http://arxiv.org/abs/2305.13661} {On the risk of
  misinformation pollution with large language models}.

\bibitem[{Prakash et~al.(2021)Prakash, Killingback, and
  Zamani}]{prakash2021incomplete}
Prafull Prakash, Julian Killingback, and Hamed Zamani. 2021.
\newblock Learning robust dense retrieval models from incomplete relevance
  labels.
\newblock In \emph{Proceedings of the 44th International ACM SIGIR Conference
  on Research and Development in Information Retrieval}.

\bibitem[{Qu et~al.(2021)Qu, Ding, Liu, Liu, Ren, Zhao, Dong, Wu, and
  Wang}]{qu2021rocketqa}
Yingqi Qu, Yuchen Ding, Jing Liu, Kai Liu, Ruiyang Ren, Wayne~Xin Zhao, Daxiang
  Dong, Hua Wu, and Haifeng Wang. 2021.
\newblock Rocketqa: An optimized training approach to dense passage retrieval
  for open-domain question answering.
\newblock In \emph{Proceedings of NAACL}.

\bibitem[{Raffel et~al.(2020)Raffel, Shazeer, Roberts, Lee, Narang, Matena,
  Zhou, Li, and Liu}]{raffel2020t5}
Colin Raffel, Noam Shazeer, Adam Roberts, Katherine Lee, Sharan Narang, Michael
  Matena, Yanqi Zhou, Wei Li, and Peter~J. Liu. 2020.
\newblock Exploring the limits of transfer learning with a unified text-to-text
  transformer.
\newblock \emph{Journal of Machine Learning Research}.

\bibitem[{Robertson and Walker(1994)}]{robertson1994bm25}
S.~E. Robertson and S.~Walker. 1994.
\newblock Some simple effective approximations to the 2-poisson model for
  probabilistic weighted retrieval.
\newblock In \emph{Proceedings of the 17th Annual International ACM SIGIR
  Conference on Research and Development in Information Retrieval}, SIGIR '94.

\bibitem[{Sachan et~al.(2021{\natexlab{a}})Sachan, Patwary, Shoeybi, Kant,
  Ping, Hamilton, and Catanzaro}]{sachan2021end}
Devendra Sachan, Mostofa Patwary, Mohammad Shoeybi, Neel Kant, Wei Ping,
  William~L Hamilton, and Bryan Catanzaro. 2021{\natexlab{a}}.
\newblock End-to-end training of neural retrievers for open-domain question
  answering.
\newblock In \emph{Proceedings of ACL-IJCNLP}.

\bibitem[{Sachan et~al.(2021{\natexlab{b}})Sachan, Reddy, Hamilton, Dyer, and
  Yogatama}]{sachan2021emdr2}
Devendra~Singh Sachan, Siva Reddy, William Hamilton, Chris Dyer, and Dani
  Yogatama. 2021{\natexlab{b}}.
\newblock End-to-end training of multi-document reader and retriever for
  open-domain question answering.
\newblock In \emph{Advances in Neural Information Processing Systems}.

\bibitem[{Santhanam et~al.(2022)Santhanam, Khattab, Saad-Falcon, Potts, and
  Zaharia}]{santhanam2022colbertv2}
Keshav Santhanam, Omar Khattab, Jon Saad-Falcon, Christopher Potts, and Matei
  Zaharia. 2022.
\newblock {C}ol{BERT}v2: Effective and efficient retrieval via lightweight late
  interaction.
\newblock In \emph{Proceedings of NAACL}.

\bibitem[{Shi et~al.(2023)Shi, Min, Yasunaga, Seo, James, Lewis, Zettlemoyer,
  and tau Yih}]{shi2023replug}
Weijia Shi, Sewon Min, Michihiro Yasunaga, Minjoon Seo, Rich James, Mike Lewis,
  Luke Zettlemoyer, and Wen tau Yih. 2023.
\newblock Replug: Retrieval-augmented black-box language models.
\newblock \emph{arXiV preprint}.

\bibitem[{Spirin and Han(2012)}]{spirin2012spam}
Nikita Spirin and Jiawei Han. 2012.
\newblock Survey on web spam detection: Principles and algorithms.
\newblock \emph{SIGKDD Explor. Newsl.}

\bibitem[{Tao et~al.(2023)Tao, Feng, Shen, Liu, Li, Geng, and
  Jiang}]{tao2023core}
Chongyang Tao, Jiazhan Feng, Tao Shen, Chang Liu, Juntao Li, Xiubo Geng, and
  Daxin Jiang. 2023.
\newblock {CORE}: Cooperative training of retriever-reranker for effective
  dialogue response selection.
\newblock In \emph{Proceedgins of ACL}.

\bibitem[{Wu et~al.(2018)Wu, Xiong, Yu, and Lin}]{wu2018memorybank}
Zhirong Wu, Yuanjun Xiong, Stella Yu, and Dahua Lin. 2018.
\newblock Unsupervised feature learning via non-parametric instance-level
  discrimination.
\newblock In \emph{Proceedings of CVPR}.

\bibitem[{Xiong et~al.(2021{\natexlab{a}})Xiong, Xiong, Li, Tang, Liu, Bennett,
  Ahmed, and Overwijk}]{xiong2020ance}
Lee Xiong, Chenyan Xiong, Ye~Li, Kwok-Fung Tang, Jialin Liu, Paul Bennett,
  Junaid Ahmed, and Arnold Overwijk. 2021{\natexlab{a}}.
\newblock Approximate nearest neighbor negative contrastive learning for dense
  text retrieval.
\newblock In \emph{Proceedings of ICLR}.

\bibitem[{Xiong et~al.(2021{\natexlab{b}})Xiong, Li, Iyer, Du, Lewis, Wang,
  Mehdad, Yih, Riedel, Kiela, and O{\u{g}}uz}]{xiong2020mdr}
Wenhan Xiong, Xiang~Lorraine Li, Srinivasan Iyer, Jingfei Du, Patrick Lewis,
  William~Yang Wang, Yashar Mehdad, Wen-tau Yih, Sebastian Riedel, Douwe Kiela,
  and Barlas O{\u{g}}uz. 2021{\natexlab{b}}.
\newblock Answering complex open-domain questions with multi-hop dense
  retrieval.
\newblock In \emph{Proceedings of ICLR}.

\bibitem[{Yang et~al.(2018)Yang, Qi, Zhang, Bengio, Cohen, Salakhutdinov, and
  Manning}]{yang2018hotpotqa}
Zhilin Yang, Peng Qi, Saizheng Zhang, Yoshua Bengio, William Cohen, Ruslan
  Salakhutdinov, and Christopher~D. Manning. 2018.
\newblock {H}otpot{QA}: A dataset for diverse, explainable multi-hop question
  answering.
\newblock In \emph{Proceedings of the 2018 Conference on Empirical Methods in
  Natural Language Processing}.

\end{thebibliography}

\appendix

\clearpage

\section{Additional Ablation Studies}


\begin{table}[t]\small
\centering
\renewcommand{\arraystretch}{1.35} 
\begin{tabular}{l c c c}
\hline
    \noalign{\hrule height 0.8pt}
    \textbf{Retriever}&Top-1 &Top-20 &MRR \\
\hline
\noalign{\hrule height0.8pt}
     DPR& & & \\ 
     ~ - 20 epochs&41.5&78.3&52.6 \\
     ~ - 40 epochs&46.6&79.7&56.0 \\
     ~ - 80 epochs&46.8&79.5&56.5 \\
\hline
     EADPR (40 epochs)&48.6&80.1&57.6 \\
\hline
\noalign{\hrule height 0.8pt}
\end{tabular}
\caption{Ablation studies on the number of training epochs. Specifically, we compare EADPR with DPR checkpoints trained over different training epochs. All models are trained using one additional BM25 negative. }

\label{tables:ablation_studies_more_iterations}
\end{table}

\begin{table}[t]\small
\centering
\renewcommand{\arraystretch}{1.35} 
\begin{tabular}{l c c c}
\hline
    \noalign{\hrule height 0.8pt}
    \textbf{Retriever}&Top-1 &Top-20 &MRR \\
\hline
\noalign{\hrule height0.8pt}
     DPR& 31.8& 74.8& 43.1 \\ 
     ~ + 1 BM25 Neg&46.6&79.7&56.0 \\
     ~ + 2 BM25 Neg&45.5&79.4&55.4 \\
\hline
     EADPR (40 epochs)&35.4&76.8&46.4 \\
     ~ + 1 BM25 Neg&48.6&80.1&57.6 \\
\hline
\noalign{\hrule height 0.8pt}
\end{tabular}
\caption{Ablation studies on the number of negatives samples used to train DPR and EADPR.}

\label{tables:ablation_studies_more_negatives}
\end{table}

\paragraph{More training iterations. }
One possible hypothesis behind the performance gain from EADPR is that the model benefits from more occurrences of positive samples during training, as EADPR uses one additional sample per instance (\ie, $p^+$ and $p^*$). To see whether the performance gain indeed comes from more training iterations of positive samples, we additionally train the baseline DPR for more epochs and measure the change in performance on NQ as the training epoch doubles. Table~\ref{tables:ablation_studies_more_iterations} shows that adding more training epochs (from 40 to 80) does not lead to significant performance gain in DPR, suggesting that the performance improvement in EADPR does not come from more training iterations of positive samples.

\paragraph{More negative samples. }
Another hypothesis is that the model benefits from more negative samples used during training (i.e. $p^-$ and $p^*$). To test this hypothesis, we compare the performance of EADPR with the baseline DPR trained using the same number of negatives per instance as EADPR. We observe in Table~\ref{tables:ablation_studies_more_negatives} that increasing the number of hard negatives used for DPR training does not increase the model performance on NQ. This is in line with the observation from \citep{karpukhin2020dpr} that DPR does not benefit much from additional hard negatives. On the other hand, we see that EADPR trained using one negative ($p^-$) and one counterfactual sample ($p^*$) outperforms DPR trained with two negative samples ($p^-$) per instance, suggesting that the performance gain in EADPR cannot be solely attributed to more negative samples used for training.

\section{Datasets}\label{appendix:dataset}

\paragraph{Single-hop QA.} All of the ODQA datasets used in this paper, \ie{ NaturalQuestions and TriviaQA}, cover Wikipedia articles written in English. Specifically, the Wikipedia corpus used in this paper is collected from English Wikipedia dump from Dec. 20, 2018, as described in \citet{karpukhin2020dpr}. Demographics of the authors do not represent any particular group of interest for both datasets. Details on the data collection can be found in \citet{kwiatkowski2019natural} and \citet{joshi2017triviaqa}.
We obtain hard negatives from the dataset provided by \citet{karpukhin2020dpr}, which is available on \href{https://github.com/facebookresearch/DPR}{https://github.com/facebookresearch/DPR}.

\paragraph{Multi-hop QA.} We train our models with the train set from \citet{yang2018hotpotqa} and evaluate them on the Wikipedia corpus of 523,332 passages. The corpus is constructed from the dump of English Wikipedia of October 1, 2017, and steps to preprocess Wikipedia documents are described in \citet{yang2018hotpotqa}. Similar to single-hop QA datasets, HotpotQA dataset does not include documents where demographics of the authors do not represent any particular group of interest.

\begin{figure}[t]
    \centering
    \begin{subfigure}[b]{0.4\textwidth}
        \centering
        \includegraphics[width=\textwidth]{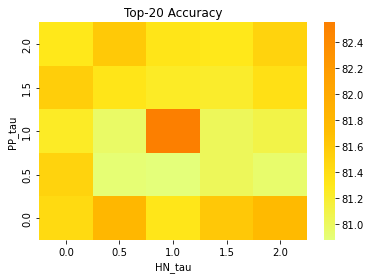}
        \caption{Top-20 accuracy}
        \label{fig:pivot-concept}
    \end{subfigure}
    \hfill 
    \begin{subfigure}[b]{0.4\textwidth}
        \centering
        \includegraphics[width=\textwidth]{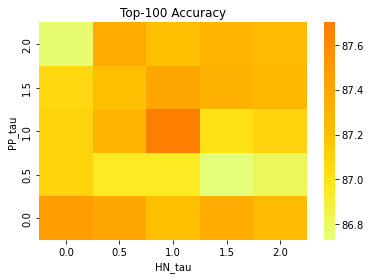}
        \caption{Top-100 accuracy}
        \label{fig:pivot-sim}
    \end{subfigure}
    \hfill
    \caption{(a) Top-20 and (b) top-100 accuracy EADPR trained on NQ with different $\tau_1$ and $\tau_2$.}
    \label{figure:hyp_tun}
\end{figure}

\begin{table*}[t]
    \renewcommand{\arraystretch}{1.15} 
    \centering
    \begin{tabular}{lcccccc}
    \hline
    \noalign{\hrule height 0.8pt}
    \multirow{2}{*}{Retriever} &  \multicolumn{2}{c}{Answer} & \multicolumn{2}{c}{Support} & \multicolumn{2}{c}{Joint} \\
    & EM & F1 & EM & F1 & EM & F1 \\
    \hline
    MDR+DPR & 49.7 & 61.0 & 41.1 & 61.5 & 30.9 & 50.2 \\
    MDR+EADPR & \textbf{54.5} & \textbf{66.1} & \textbf{47.1} & \textbf{68.2} & \textbf{35.5} & \textbf{56.4} \\
    \hline
    \noalign{\hrule height 0.8pt}
    \end{tabular}
    \caption{Reader performance on HotpotQA dev set. The QA performance is measure based on Exact Match (EM) and F1 scores of answers (Answer EM/F1), supporting sentences (Support EM/F1), and both (Joint EM/F1).}
    \label{tab:mdr_QA_full}
\end{table*}

\section{Implementation Details}\label{sec:implementation}
\paragraph{Dense Retrievers. }Our implementations of dense retrievers follow the dual encoder framework of DPR~\citep{karpukhin2020dpr}, where each encoder adopts BERT-base~\citep{devlin2019bert} (110M parameters) as the base architecture.
For experiments on ODQA benchmarks in Section~\ref{sec:exp.2}, we train all implemented models for 40 epochs on a single server with two 16-core Intel(R) Xeon(R) Gold 6226R CPUs, a 264GB RAM, and 8 24GB GPUs. For \our~training, we set batch size as 16, learning rate as 2e-5, and eps and betas of the adam optimizer as 1e-8 and (0.9, 0.999), respectively. Note that we conduct experiments on the NQ and TriviaQA benchmarks under the same settings used in \citet{karpukhin2020dpr}. 
Among the hyperparameters $\{0.1,0.2,0.5,0.9,1.0\}$, we choose 1.0 for the balancing coefficient $\lambda$ for counterfactual samples in Equation~\ref{eq:dpr}. The weight hyperparameters $\tau _1, \tau _2$ in Equation~\ref{eq:final} are set as $1.0$. We find the best hyperparameters for $\tau _1, \tau _2$ using grid search. Figure~\ref{figure:hyp_tun} shows the performance of EADPR trained with different combinations of $\tau _1, \tau _2$.

\paragraph{Readers. }For reader in single-hop QA experiments, we consider two models: 1) the extractive reader from \citet{karpukhin2020dpr} implemented on pretrained BERT models~\citep{devlin2019bert} and 2) Fusion-in-Decoder reader~\citep{izacard2021fid} based on pretrained T5-base~\citep{raffel2020t5} models. We conduct inference for the reader on a single 24GB GPU with the batch size of 8. For all experiments, we conducted a single run of each model tested. Our empirical findings showed little variance in the results over multiple runs. 

For reader in multi-hop QA experiments, we use the extractive ELECTRA~\citep{clark2020electra} reader provided in \citet{xiong2020mdr}. Reader inference is conducted on a single 24GB GPU with the number of input contexts limited to 20. For all experiments, we conducted a single run of each model tested. Our empirical findings showed little variance in the results over multiple runs.

\paragraph{Multihop Dense Retrieval. } The classic approaches to multi-hop QA usually involve decomposing questions into multiple subquestions, retrieving relevant contexts for each subquestion, and aggregating multiple contexts into a reasoning path \citep{asai2020learning}. In line with these studies, \citet{xiong2020mdr} train a Multihop Dense Retrieval (MDR) to construct reasoning paths by performing dense retrieval in multiple hops, each time with query representations augmented using the retrieved passages. MDR is paired with a reader that takes reasoning paths as inputs, and the QA performance is measured based on Exact Match (EM) and F1 scores of answers (Answer EM/F1), supporting sentences (Support EM/F1), and both (Joint EM/F1).

We implement MDR using EADPR following \citet{xiong2020mdr} but with some constraints due to limited computing resources: (1) we train our models on smaller batch sizes of 120 compared to 150 in the original paper; (2) our MDR implementation is not optimized using the memory bank mechanism~\citep{wu2018memorybank}; (3) we generate 20 candidate reasoning paths (\ie, beams) instead of 100 in the original paper. Table~\ref{tab:mdr_QA_full} reports in detail the QA performance of the reader when paired with different MDR.

\paragraph{Software Packages. }We use NLTK~\citep{bird2009natural}~\footnote{\url{https://www.nltk.org/}} and SpaCy~\footnote{\url{https://spacy.io/}} for text preprocessing. Following DPR, we adopt FAISS~\citep{johnson2019billion}, an approximate nearest neighbor (ANN) indexing library for efficient search, in our implementation of \our. DPR also uses an open-sourced library for logging and configuration named Hydra~\footnote{\url{https://github.com/facebookresearch/hydra}}, which we use to configure our experiments. No modification has been made to the aforementioned packages.

\paragraph{Terms and License. } Our implementation of \our~is based on the public repository of DPR~\footnote{\url{https://github.com/facebookresearch/DPR}}, which is licensed under Creative Commons by CC-BY-NC 4.0. The indexing library FAISS is licensed by MIT license. Both ODQA datasets, NaturalQuestions and TriviaQA, are licensed under Apache License, Version 2.0. We have confirmed that all of the artifacts used in this paper are available for non-commercial, scientific use.

\end{document}